\documentclass{article}
\usepackage[utf8]{inputenc}

\usepackage[inner=25mm,outer=25mm]{geometry} 
\usepackage{times}
\usepackage{graphicx}
\usepackage{natbib}
\setcitestyle{authoryear,open={(},close={)}} 

\title{On the Computational Modeling of Meaning: \\ Embodied Cognition Intertwined with Emotion }
\author{Casey Kennington \\ Department of Computer Science \\ Boise State University}
\date{}

\begin{document}

\maketitle

\begin{abstract}
This document chronicles this author's attempt to explore how words come to mean what they do, with a particular focus on child language acquisition and what that means for models of language understanding.\footnote{I say \emph{historical} because I synthesize the ideas based on when I discovered them and how those ideas influenced my later thinking.} I explain the setting for child language learning, how embodiment---being able to perceive and enact in the world, including knowledge of concrete and abstract concepts---is crucial, and how emotion and cognition relate to each other and the language learning process. I end with what I think are some of the requirements for a language-learning agent that learns language in a setting similar to that of children. This paper can act as a potential guide for ongoing and future work in modeling language. 
\end{abstract}

\section{Introduction}

\emph{How can machines understand language?} is a question that many have asked, and represents an important facet of artificial intelligence. Large language models like ChatGPT seem to understand language, but as has been pointed out \citep{Bender2020-ck,Bisk2020-wy}, even large, powerful language models trained on huge amounts of data are likely missing key information to allow them to reach the depth of understanding that humans have. What information are they missing, and, perhaps more importantly, what information do they have that enables them to understand, to the degree that they do? Current computational models of semantic meaning can be broken down into three paradigms:

\begin{itemize}
    \item distributional paradigms where meaning is derived from how words are used in text (i.e., the notion that the meaning of a word depends on the ``company it keeps," following \cite{Firth1957-xy})
    \item meaningfulness of language lies in the fact that it is about the world \citep{Dahlgren1976-oh} and grounded paradigms are where aspects of the physical world are linked to language (i.e., the symbol grounding problem following \cite{Harnad1990-fr})
    \item formal paradigms where meaning is a logical form (e.g., first order logic as in \cite{LTF_Gamut1991-ma})
\end{itemize}

\begin{figure*}[t]
    \centering
    \includegraphics[width=.9\textwidth]{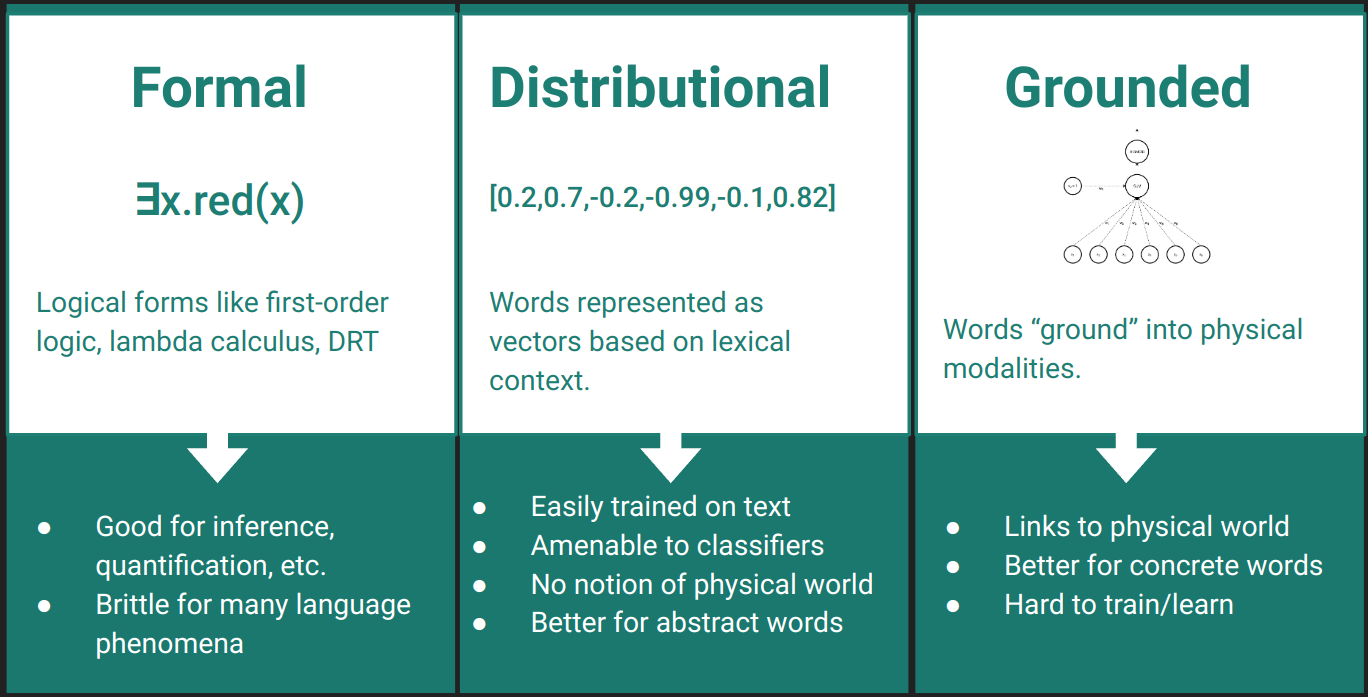}
    \caption{Three approaches of computational semantics.}
    \label{fig:semantics}
\end{figure*}

Figure~\ref{fig:semantics} shows examples of the three paradigms to computational semantics and the kinds of language phenomena that they model well. These paradigms to computational semantics have been applied in various models that represent remarkable progress in recent years. However, now that large language models and other AI models are more widely used, it is clear that there are limits to their `understanding` (if they fully understand, then why is prompt engineering necessary?) which has prompted some to claim that a full, unified model of computational semantics is only possible if it goes through the same language acquisition process that children do. 

Even if computational models of language meaning do not need to learn in the same settings and progression that children do, it is useful to make an appeal to what is known about how children do learn language in order to guide current and future modeling efforts to enable models to have a more holistic understanding of language.\footnote{Why? Because understanding and misunderstanding is a vexing societal problem and a scientific understanding of how to acquire, represent, and apply language in a computational model will tell us something about what language is, which may help us overcome those vexing misunderstandings.}

This paper represents such an appeal. At least it represents this author's attempt in the past decade to synthesize what is known about child language acquisition and model semantics more holistically. 

\section{Why do Children Speak?}
 
One goal of my research is to determine the \emph{setting} and \emph{requirements} for language learning; specifically I have been searching for environmental reinforcement signals that a child could use to know that they were aligning with language speakers, and what the parameters of the pre-linguistic (i.e., before the first real words---not just babbling---are uttered with some kind of intent) language learner might be. 

In ``Child's Talk: Learning to Learn Language" \citep{Bruner1983-uv}, we get at some important basics (the following are quotes from Bruner): it is banal to say that infants (and, generally, humans) are social; they are geared to respond to the human face, voice, action, and gesture. Children seem to want to coordinate their actions with, or at the very least mimic the behavior that they see in conspecific entities: i.e.,  with their mothers. \cite{Snow1977-zp} noted that children have basic needs that might contribute to spoken interaction, namely that children aspire to affection and intimacy with their caregivers. Mothers are able to track a child's progress and act accordingly. 

Mothers seem to follow the Gricean maxims of quantity, quality, relation, and manner \citep{Grice1981-ad}. The initial cognitive endowment appears to be that it is goal-driven activity, is social and communicative---self-propelled and self-rewarding---constrained, ordered, systematic, familiar, often referring, and surprisingly abstract (as opposed to \emph{concrete}, which is usually what is assumed when considering that children first refer to physical objects). 

A book more directly related to what I was looking for was \emph{How Children Learn to Learn Language} by Lorraine McCune (\citeyear{McCune2008-oy}), who claims that the basis of meaning is grounded in embodiment, something I had not really before considered. This way of thinking was quite different from the general NLP thesis that meaning can be derived from text alone, with word embeddings dominating the ``semantic" side of the field. Text isn't embodied, and children don't learn language via the medium of text. In fact, if words are to be learned, then children must attend to physical objects (including self and others), and one thing that makes objects salient to children is the fact that they move \citep{Hollich2000-rt}. 

McCune makes the case (synthesizing other work) that linguistic patterns and order emerge not necessarily with explicit instructions---language learning doesn't requite a curriculum, though caregivers use simple words and phrases at first. Children are interested in novelty which gives them sensitivities in information coming through all of the sensory inputs and internally (e.g., proprioperception) from their own bodies. McCune noted four stages that children go through as they learn language (p.27):

\begin{enumerate}
    \item Sensitivity to expressiveness (i.e., movement and sound)
    \item Transcendence of expressive qualities and knowing attitude (i.e., the child recognizes that actions and sounds are communicative)
    \item Denotative reference and semantics correspondence (i.e., children begin to refer to objects and learn their designations)
    \item Shared perceptual and representational settings (i.e., children learn language in a shared space where they and caregivers directly perceive objects and each other)
\end{enumerate}

To learn language, eyes are important; children can follow paternal line of regard (i.e., triangulate what another person is looking at) by just seven months. Children somehow know that the eyes are an important modality of attention and they wonder why the eyes of others aren't pointed at themselves because the attention of others is something that children seem to innately seek. Reference to visually present objects (the subject of my PhD dissertation) is an important step in the developmental process, which coincides with the development of theory-of-mind (i.e., the child comes to the realization that they are an individual separate and distinct from others, and they allow others the same endowment of distinct individual identity and frame of reference to the world). But reference doesn't come first; there are other pre-linguistic parameters that must be in place before reference to visual objects can begin. 

My primary takeaway from McCune's work was that children are motivated, intrinsically, to \emph{interact with other human beings} and that language learning likely would not take place without interaction, nor without the motivation to interact. Other work reinforces spoken interaction \citep{Elman1998-nq}, and another reinforces that there is no overt supervision signal; children just need to explore and observe to find patterns and regularities, and once a regularity is learned, exploit it to learn more abstract regularities \citep{Eimas1987-ef}.

\subsection{Nature or Nurture?}

Part of my quest to understand the settings and parameters for language learning has meant taking a stance on the nature (à la Chomsky) vs.\ nurture debate (spoiler alert: both are required, but with some nuance). It is clear that there is some degree of cognitive scaffolding that uniquely affords humans the ability to think and talk about abstract ideas using speech and other communicative mediums. Furthermore, it is known that pre-linguistic infants possess ``highly developed perceptual mechanisms for the perception of speech" \cite{Eimas1987-ef}. It is also clear that a fairly substantial degree of linguistic exposure is necessary for children to learn language, and by language I don't just mean syntax. Important to this debate was an observation made in \cite{Marcus2004-ia} that learning is not the antithesis of innateness, but one of its important products. \cite{Tomasello2009-wq} makes a strong case that language requires experience and that languages are socially constructed even between the child language learner and the parent. 

It is known that comprehension of speech occurs ahead of production of speech, and that visual, physical context is critical to learning language. An important takeaway from that book is that adults are not simply performing random behaviors, they are performing intentful behaviors, and children pick up on those intents. Understanding intent first seems to be a precursor to language. 

For example, a child who is old enough to make use of hands to reach objects understands the intent to make an effort to reach for an object, so when that child sees another person reaching for an object there is an understanding of intent behind that other person (an important part of theory-of-mind); i.e., the person wants to grab the object. Sounds that come from human mouths that accompany those kinds of actions form the basis for language because \emph{language builds on understanding of intent}. That, once again, simply means that the child learner requires a body to make utterances, to enact intentful behaviors, and to experience them personally in order to recognize them in others.

\section{Modeling Challenges}

If embodiment is necessary, does it matter what kind of body? Searching for an embodied setting led me to explore robots as a body for my computational model, but I had no experience with robots. I did not want to build one. I was, however, an interested consumer that wanted to put my incremental (i.e., processes at the word level) dialogue system onto a robot platform so the physical interaction could take place. I opted to purchase several Anki Cozmo robots because they were affordable, small, and had a Python SDK that gave me access to sensors and control over actions. 

Learning how to use the robot took longer than it should have, requiring branching into the field of human-robot interaction (HRI) because we had to establish that the Cozmo robot was the right one for the job of first language learning, that people would actually treat it like a child and did not have adult-level cognitive capabilities \citep{Plane2018-wh}. There were technical challenges in this regard; getting our spoken dialogue systems to play nicely with the robot SDK took a lot of effort and we knew that the model of semantics that we had espoused wasn't quite right to work well with the robot's sensors. The importance of good technical infrastructure cannot be understated, yet it takes up a lot of time because without it we cannot do productive research. 

\subsection{Objects in the World}

\emph{How Children Learn the Meanings of Words} by Paul Bloom \citep{Bloom2000-pj} focuses somewhat more on the child who was ready to learn words and what those words might mean. Some words refer to objects, so what do children assume about those objects? Bloom mentioned Spekle objects with four important principles:


\begin{itemize}
    \item cohesion --- objects hold together
    \item continuity --- objects remain even if they disappear from view
    \item solidity --- objects are solid
    \item contact --- objects can interact with each other including people
\end{itemize}

This concept is important because children interact with objects and people before they begin referring to objects using words, which requires that they have some kind of understanding of those objects; at least how they feel, their potential affordances (e.g., a ball is roll-able, a box can hold objects) which is what is potentially grounded into when the children begin to learn reference words. This puts reference (and affordances) in a central position to meaning, at least when children are learning words that refer to concrete, physical objects. Bloom states that if reference is central to meaning, then meaning is not determined by mental representations. This is an important point because that affects the model (whatever a mental representation is). 

Similar to Bloom, \cite{Westermann2017-da} looked at the literature on early word learning in children. Children learn words slowly; if children could learn words quickly, we would not see a strong correlation between how much parents talk to their young children and a child's vocabulary scores, but we do (nod to Tomasello). There is such a thing as fast-mapping in older (though still young) children, but the initial words are not so easy. 

Production (i.e., how a word is used / uttered) is the ultimate demonstration that a child has learned a word. Interaction requires speech, and speech unfolds phonetically over time, so listeners must interpret words incrementally, one syllable at a time. \cite{Jaffe2001-pm} is an entire book on this subject with the thesis that timing posits (rather, builds upon) the notion that the give-and-take and timing of that give-and-take are foundational before other cognitive development can take place. Not just spoken language; two objects can't occupy the same space so there is a give-and-take in the use of spaces and a give-and-take in the use of things like speech as a communication medium because we can only attend to one voice at a time, and because speech is manifested as compressed air within a specific space, so only one person can talk at a time if anyone is to be understood. This is either a handy thing because human attention is limited to one thing at a time, or it might actually be one of the things that limits human attention to one thing at a time. 

Relatedly, one important point that is the main thesis of \cite{Mix2010-ll} is that our cognitive functions are housed in a body that lives in time and a kind of space with specific degrees of freedom (e.g., three axes of movement are allowed in that space) and our cognitive machinery builds its understanding (and as Johnson and Lakoff observed, metaphors \citep{Lakoff2008-ud}) upon the spatial foundation of language and cognition. 

Imagine communicating without prepositions, or the information encoded in action verbs that connote activities that occur in time and space. This is where mothers shine: mothers take note of the space-time constraints then have simulated dialogues with their babies; when a child doesn't respond to a turn, the mother still allows the duration of what might have been a turn to elapse before taking the dialogue floor again. Mothers' high responsivity  facilitates their infants' cognitive development. This is the literal cradle of social adaptation; cognition and cognitive development are inseparable from social adaptation---but it must be interactive in that cognitive beings are participating in the social interaction. 

Digging deeper into child psychology, \cite{Hetherington1999-lv} explain a few important things. First, that parents repeat what children say, they don't overlap their speech with children (i.e., the dialogue has very easy-to-distinguish turns), the speech is simplified with primary content at the end of what is being said, and parents repeat what children mean by rephrasing in a grammatically correct way. Thus parents assume the child has an egocentric frame of reference of the world (i.e., they can only take on their own frame of reference---they haven't discovered that others have their own frame of reference). Parents keep a level of complexity just ahead the child which gives the child enough novelty, thereby holding attention and learning. 

Taken together, physical, co-located interaction between parent and child is key. Children are motivated to interact, and caregivers assume an egocentric frame of reference for the children, meaning that parents don't refer to the objects, they often name the objects that the children are already attending. Learning that was helpful for me because those are some of the parameters that need to be in place for a computational model: 
\begin{itemize}
    \item it must be in an interactive setting
    \item the learner should probably be embodied (at the very least it should have sensory input)
    \item because of the ego-centricity we could assume that when objects are referred, it is because they are already salient to the child
\end{itemize}

Which computational model can fulfill those requirements?

\subsection{Deep Learning and Transformer Language Models}

Deep neural networks are the mainstay of most NLP tasks and the latest architectures of the time led to a new language model that dramatically altered everything. \cite{Vaswani2017-kv} introduced the attention mechanism and \cite{Devlin2018-lz} made attention and transformer architectures work in NLP as a new way to use a pre-trained language model called BERT to do anything. The caveat was that it was trained on large amounts of \emph{text}. Broad-sweeping claims followed: BERT and more powerful derivatives were at the basis for artificial general intelligence, etc., etc. That caused me to raise an eyebrow because throwing text from books and websites at a model and using a learning regime of guess-the-word within a sentence wasn't anywhere close to how children learn language, if all of those books I had been reading about child language learning (and my own experience with my children) were to be believed. 

But do we really need to be concerned with mimicking exactly how humans learn language? After all, airplanes fly without flapping their wings. Two responses to that: first, the reason deep learning works so well is \emph{because} it is bio-inspired, so there is something potentially useful about trying to mimic biological processes. Second, language is an ability that is so uniquely human that understanding it means understanding how humans acquire and use language. 

Thankfully, I wasn't the only one who had reservations about BERT and derivatives thereof. \cite{Bender2020-ck} highlights some of the important reservations that many in the field have with assigning so much meaning and understanding to BERT-like transformer language models. Others have followed with their own skepticisms since phrases like  \emph{large language models} like \emph{ChatGPT} became part of everyday vernacular. 

\section{Meaning and Embodiment}

\subsection{What is Context?}

What possibly annoyed me most in my investigations was the claim that the language models were following a Wittgenstein view of meaning in language, that meaning is derived from how it is used in a context. What context? The assumption is lexical context; i.e., words used in the textual context of other words. But there is also physical context, and I believe that is more likely what Wittgenstein meant. I picked up his Philosophical Investigations \citep{Wittgenstein2010-ji} in 2019 and what luck our library had it in English and its original German. I read both in tandem, and while I found the translation reliable (mostly), it bothered me that the accepted interpretation of Wittgenstein's stance on language meaning was text-centric, or at least context only meant what was spoken or written previously. This was late Wittgenstein when he thought he had settled language as a more formal system, but then spent time with children and (I conjecture) realized that the child mind is not the same as the adult mind.

More evidence that he meant that meaning comes from physical context: ``There is a gulf between an order and its execution. It has to be filled by the act of understanding" (1.431) and not disconnected from the body (1.339). I interpret this to mean that meaning requires action, or a body to act in, because meaning is grounded in bodily movement. The word \emph{throw}, for example, isn't just an idea and it's not just something we see someone else do, we have muscle memory or throwing that is part of the meaning of throw.  He also brings up color and shape (1.72-74), that words refer to objects which themselves have affordances (1.11), and mentions that language use is first in reference to deictic (i.e. pointing) gestures. In other words, \emph{at first, language is grounded in the physical world}. Only after a conceptual foundation of concrete concepts do we get to abstract language games (1.270+) and thinking about thought (1.428); i.e., use language to construct meanings of abstract words and abstract thought only after concrete scaffolding. In contrast, language models were only focusing on the last bit: use a language game of ``guess the word that I randomly covered" to think about abstract thought, but a model distributed in text, \emph{all} words are treated as if they are abstract. This idea was hard to convey in some of my (rejected) papers. Wittgenstein did explore how words come to arrive at a shared meeting between speakers that don't have observable thoughts, which is what language games are for, an observation explored deeply in \cite{Clark1996-zu}. 

In any case, language models are here and making an enormous splash and winning all of the benchmarks. If you aren't using language models in your research then at least one reviewer will use that as a justification for rejecting your paper. That's not to say that transformer language models don't have merit---they really do---but try using one out of the box on a robot, show the robot an apple, then ask if the apple is red.\footnote{Though there is now a trend in multimodal language models that at least bring the visual modality into language, see \cite{Fields2023-zv} for a review.} The language model doesn't know anything about redness, it only knows that red is a color and might be able to list of some objects that are typically red. That is changing with visual and other multimodal language models, but as observed by \cite{Schlangen2023-sc}, the language ``learning" progression made in the NLP community starting with transformer language models then working towards more embodied notions of meaning is the opposite direction of how children learn language. Clearly there is top-down processing that happens in cognitive processes and they are in play early on in a child's life, but large language models were completely lacking anything bottom up from the pyhsical world. 

\subsection{Embodied Cognition}


Johnson, along with George Lakoff, has been an early proponent of embodied cognition and carries forward the research of the time in his later book \citep{Johnson2008-zd}.  \cite{Di_Paolo2018-ba} puts language and bodies together. Both make a strong claim that the fact that our bodies are unique and distinct from other bodies, allowing interaction to take place, and putting language at the social level between linguistic bodies. Moreover, the categorical gap between sensorimotor life and the life of language is not only big, it is largely uncharted scientifically. 

We cannot separate bodies from what they \emph{do}, making people with bodies agents (in that they act), where agency is the active regulation of tensions between different negative tendencies; the actions of the agent are guided by positive norms that emerge dialectically out of opposing negative ones. Di Paolo of course mentions reference and social interaction, but the main thrust is that without the precarious materiality of bodies, there would be no meaning and no minds (p.110). 

The idea that bodies are important to meaning is not new. \cite{Brooks2003-ox} predicted what the world of AI might look like in the future (i.e., today, 20 years after its publication) and embodiment was not out of the equation. Brooks mentions Kismet, a simple robot that could respond to stimuli in ways that humans interpreted as somewhat intelligible. But, as the author admits, what Kismet cannot do is actually \emph{understand} what is said to it. Nor can it say anything meaningful, and it turns out that neither of these restrictions seem to be much of an impediment to good conversation (sorry, chatbots). 

Moreover, according to Brooks, researchers are operating in an underconstrained environment, and as they follow up interesting research ideas, they are tempted---and succumb---to make their abstract world more interesting for their research ideas, rather than being faithful to the reality of the physical world. This is exactly the issue I take with language models that are perfectly satisfied with deriving meaning through abstract text--partly because the datasets are easier to come by than the painful collection of often stilted spoken dialogues accompanied by a recording of the physical environment in which the spoken dialogue took place.

Embodied cognition is not without its critics, and there are plenty of theories of cognition that don't require a body. \cite{Baggio2018-lp} in \emph{Meaning in the Brain} takes a step back and looks at meaning from a different perspective: meaning is not a given, but rather the result of a constructive process that uses knowledge to make sense of sensory signals. So there is sense information, but the mechanisms in the brain aren't just reading sensory input in and finding patterns; rather, the brain is actively trying to find meaning. That's partly because prediction of possible pathways of conversation is a fundamental process of what the brain is doing during conversation, and that often drives the meaning of a given situation, linguistic or not. Embodied perceptual activations are not required for representing input's meaning, which is what makes language so ultimately useful. That of course only counts at the adult level, but each language learner needs to arrive at that point individually. Baggio does take that into consideration, noting that first words, in particular common nouns and their meanings, are learned in the first year of life, in social contexts where coordination between the infant and caregiver is generally the primary goal of interaction. 

In the first two to three years of life, children learn largely by observing or imitating adults. Furthermore, considering the world outside just a single brain, Baggio quotes Miller's Law: In order to understand what another person is saying, you must assume it is true and try to imagine what it could be true of (Grice's maxims apply here, mostly the maxim of quality). Children must do that or they might not learn to speak at all. They only learn later about lies and manipulation---unfortunate aspects of the human condition---but no one would learn language if children assumed a-priori that nothing was true. 


\subsection{Which Body?}

If embodiment is required for holistic computational modeling of language, then which body? Virtual agents offer a kind of body that could be an important stepping stone. They can be made to look human, which may also be important. The main drawback for me in my quest for holistic semantics was the fact that virtual agents exist in a virtual world. What is required is a body that can enact in the world that humans share with each other. That left robots. 

\cite{Cangelosi2015-jl} and more recently \cite{Lee2020-yb} bring some clarity to the stance that embodiment is crucial to cognitive development of minds. Like humans, robots are a kind of body that don't just observe the world; a robot is an autonomous physical device, of any shape or form, that can sense and perform actions in its working environment \citep{Lee2020-yb} (p.20). That means that the robot has to be able to act, at least to some degree, where it is physically located. Humans have the same limitation, though of course humans can control things remotely using technology, but our own limbs are limited to what they can reach here and now. 

\cite{Lee2020-yb} makes the case that the Turing Test is not a proper test of intelligence because words and concepts, including the most abstract, must have their meanings ultimately grounded in sensory-motor experience. That's not a reductionist account that all meaning is eventually grounded; it's more of an account of a proper progression: one cannot come to learn the meaning (connotation) of abstract ideas like \emph{democracy} without the vocabulary required to define democracy, and so on until one reaches the point where words are not learned by other words. Rather, they are concrete words that denote physical objects or object attributes.  \cite{Vincent-Lamarre2016-gi} showed how recursively considering words that define other words in a dictionary eventually lead to a core subset of words that all other words are defined upon. 

Modern robotics has shown how important embodiment is \citep{Prescott2023-pd}. Without a sensory-rich body, perception (as we know it) is impossible. And enaction (i.e., acting intentfully and not just sensing) is also vital---our actions are entangled with our thoughts just as much as perception. The life process, the life cycle of the individual, cannot be separated from embodied cognition. This is the difference between biological brains and computer brains. Indeed, \cite{Clark2013-rs}, required reading for anyone interested in first language acquisition, mentions that children seem to act out things as they are learning words, like opening and closing doors as they learn words \emph{open} and \emph{close}. Transformer models definitely don't do that, resulting in a meaningful lack of meaning.

\section{Emotion and Language}

Though not strictly a book about child language development, \cite{Alan_Sroufe2009-wa} reported a longitudinal study of a group of people across decades to track their development from birth to adulthood that gives a broader picture of how humans develop in an individual, familial, and societal context. One of the main themes of the book is behavior (since behavior is something that can be observed), and what behavior means to the organization of an individual. How does this relate to language? 

Central aspects of individual organization originate in the organization of early relationship. Language is part of that organization since it is a method of communication that maintains, fosters, or harms those relationships. Another is the main theoretical thrust of the book: that organization is the fundamental feature of behavior---i.e., language is part of the organizational structure itself. Organization is revealed in the interplay of emotion, cognition, and social behavior; development is defined by changes in organization of behavior over time, and organization of behavior is central to defining individual differences.

If development of an individual human means that they are organizing their behaviors (emotion, cognition, and social behavior and the interplay between emotion, cognition, and social behavior) then language development---which is an organizing behavior---plays a central role in the organization of emotion, cognition, and social behavior. 

The idea that language plays a role in the organization of social behavior was clearly laid out in \cite{Clark1996-zu}, along with its accompanying idea that social behavior in turn plays a role in the organization of language because people have to coordinate what they mean when they speak with each other. Furthermore, that cognition plays a role in language and that language plays a role in cognition is well-established---language and cognition are often considered one and the same. But what about emotion? If emotion plays a role in the organization of behavior, and if emotion has a tight interplay with cognition and social behavior, what does emotion have to do with language?

Like many, I considered emotion to be part of the human experience, but clearly separate and distinct from cognition.\footnote{This is perhaps in part due to my affinity for Commander Data on \emph{Star Trek: The Next Generation} who was a conscious and highly intelligent android, yet emotionless.} In fact, emotion was in my view often a hindrance to true linguistic understanding because emotion colors understanding in potentially the ``wrong" way. The more we could separate emotion from meaning the better. That researchers were trying to model the ability to recover emotional content from text (e.g., a short post on a social media site) did seem useful, but the goal there was utilitarian, not to uncover the meaning of language. 

My stance on emotion began to change when I took on a master's student, David, to whom I handed the Cozmo robot and tasked him with putting everything together we knew about language, meaning, spoken dialogue, and placing the robot in a setting where it could learn language from people without knowing any language a priori. After considering the task, he asked a question I had not considered (students tend to do that): ``If we bring in people to talk to Cozmo, why would they care to help it learn language?" I don't think I grasped the question fully at the time. It partially meant that the science we were trying to advance was different from other science in that we weren't just observing people behaving, we were asking them to behave in a way such that they might help this little robot learn some words; i.e., we were asking them to be caregivers for an hour. David was concerned that paying participants for an hour of their time wasn't enough because there was no connection between them and the robot to care that the robot actually learned anything. 

All of the literature I had read up until that point about what mothers do to foster development, particularly language development, backed up that concern. The robot had no mother. What we possibly lacked from the participants was ``buy-in" to the needs of the robot to learn language. One way to potentially convince people to buy-in was to make the robot display behaviors that would motivate the participants to buy-in in a way that capitalized on a general human decency to help others.\footnote{The ethical implications of this were always a concern to us.} But the displays had to be age-appropriate for the robot; it was supposed to be like a pre-lingusitic child, after all, and what kinds of behaviors can pre-linguistic children engage in to capitalize on the decency of others to help? Well, children smile and they cry---they display emotion. David put in the due-diligence in the relevant literature and found that a number of important behaviors had emotional underpinnings that could help facilitate language learning; for example, \emph{confusion} and \emph{curiosity}. 

That put us in a difficult position that we had been in before with human perceptions of the robot's age and cognitive level: what behaviors could we make the robot do in order to make people think that the robot was displaying some kind of emotional state? With Cozmo we struck gold because Cozmo had nearly 1,000 short, pre-defined behaviors that we could easily invoke and some of them were designed to have emotion content (e.g., the robot smiles, makes a ``happy" noise, and moves its lift was meant to display happiness). David painstakingly video/audio-recorded all of the behaviors and put the recordings on a crowd-sourcing website, asking people to rate the behaviors for their emotional qualities. That work led to a model of emotion recognition, not from humans (!) but inferring what emotion people would attribute to a robot behavior based on the behavior itself (i.e., the movements, face, and sounds from the robot). That led to \cite{McNeill2019-ia}, which was only the beginning of what we thought was a temporary, minor detour down the path of emotion and what it might have to do with language. 

\subsection{Concrete Affect, Abstract Emotion}

My original hypothesis about emotion and language was that because emotion exists and is used to display information about the state of a child before language is learned, then emotion must be something that, like other perceptual modalities, language grounds into (i.e., the modality itself is part of the meaning) as language is learned. To find out more about emotion in general (not just how it relates to language), I picked up \cite{Lane2002-pk}, a dense but thorough read, and had to step back---way back---from what I thought I understood about emotion. That led me to read a few papers on the subject of language and emotion with a more open mind; these were neurosicence and psychology papers (the NLP community was only interested in how to infer the emotion of the writer of a piece of text, not in how it relates to meaning). 

Two papers, both originating from Gabriella Vigliocco, really changed my understanding because they made a strong case that abstract words are more directly tied to emotion than concrete words \citep{Vigliocco2014-km,Ponari2018-ve}. This agreed, it seemed, with \cite{Barrett2017-nu} which synthesizes the latest emotion research. At the very least, I came to the understanding of the difference between emotion and affect (most people use the two terms interchangeably) in that emotion is tied to the linguistic and cognitive system making emotions to a large degree socially constructed, whereas affect is more basic; grounded in embodiment. 


\subsection{Emotion and Cognition}

\emph{A Child's Path to Spoken Langage} by John L. Lock \citep{Locke1995-tz} makes some important arguments vis-à-vis cognition and emotion. As has been noted by others (some of them citing Locke's work), access to the vocalizations of one's species is simply not enough. Furthermore, an inclination to imitate and all the good communicative intentions in the world are insufficient for being a speaker of a language. Rather, the availability of appropriately interactive tutors are part of the story. That means putting an agent (or computational model) in a place where it can observe language, be it text or even referring expressions made to visually present objects, does not bring the child to language capabilities as much as \emph{participatory} interaction. But interaction doesn't take place with just anyone---language is developmentally additive in the sense that the multiplicity of cues that trip off phonetic categories are piled on top of the prosodic, affective, and speaker-identifying cues that form the infralinguistic (i.e., non-linguistic) core of our vocal messages. It seems that infants need to know and have experience with the identity and intentions of those who are speaking to them. This could be because mothers imitate children 90\% of the time, not the other way around. 

Reinforcing some of the themes mentioned above, Locke further argues that language development proceeds from the general to the specific, and complex structures evolve by differentiation of a larger entity into smaller parts or functions (words to phonemes, phrases with prosody to words). From its earliest opportunity, the infant seeks out the particular kinds of stimulation that it enjoys and that its brain may need in order to develop maximally. Young humans express more interest in the eyes than in any other region of the face, and it seems that the human infant is largely preadapted to indexical and affective communication. Even little monkeys and apes do not ``while away the hours in idle vocalization;" quite the contrary, but little humans babble. 

An anecdote illustrates this. When my daughter was learning to speak, her word for all non-flying animals was \emph{cow} because we lived near a farm and she saw cows a lot. Only later did she use size to distinguish between big animals and small ones, thus the concept \emph{dog} emerged for the latter category. As she learned more words for more specific species, she picked up on the details that distinguished them. 

Studies reveal that it is not just the sound of speech that sets infants to vocalizing or reinforces them for doing so: the person doing the speaking must be physically present and it may help if the speaker is visibly looking at the child; vocal imitation may occur more commonly when the baby can see the person who is talking or see a person while there is talking. It's worth noting research that sorts children into two cognitive camps: some children are more referential so their first words largely refer to physical objects, while other children are more expressive so their first words are more egocentric about their own feelings and needs. 

In \cite{Locke1995-tz}, an entire chapter is dedicated to emotion and language development. Effectance (i.e., motivation to act and interact) and affect play several major roles in human communication. Certainly they expand the infant's capacity for intelligent behavior by pushing it to explore and to interact with the people and objects in its environment. Affective displays provide parents with a basis for social responding and cues they may use in adjusting the psychological and physical care of their infant. The experience of emotion fills infants with energies that are dissipated by behaviors (such as squealing), which are by their very nature communicative (p.328). Piaget said that affect plays an essential role in the functioning of intelligence. Without affect there would be no interest, no need, no motivation; and, consequently, questions or problems would never be posed, and there would be no intelligence. Affectivity is a necessary condition in the constitution of intelligence. 

Locke also cites Sroufe and Waters who worked on the longitudinal study mentioned above \citep{Alan_Sroufe2009-wa} early on in the longitudinal project where they note that that cognitive advances ``promote exploration, social development, and the differentiation of affect; and affective-social growth leads cognitive development [...] neither the cognitive nor the affective system can be considered dominant or more basic than the other; they are inseperable manifestations of the same integrated process [...] It is as valid to say that cognition is in the service of affect as to say that affect reflects cognitive processes." Moreover, in the real speech of sophisticated speakers, where both linguistic content and vocal affect are present, one type of cue does not preempt the other, and for speech to work this must be the case. 

Listeners must know both what the speaker is saying and what he intends by saying it. Speakers duplexly pick up information about the linguistic content and the speaker affect because the cues to these things are of different sorts and are processed by different brain mechanisms. Thus, according to Locke, the meaning of an utterance is in the linguistic content, but the intent of the speaker who made the utterance is in the affect and emotion. In fact, children are adept at reading intents of others via affect and emotion, before they can even speak or really understand words. 

\subsection{Modeling Emotion}

Taken together, the above discussion means that the separation of language from emotion in computational models is going to lead to something that is only an approximation of what a language model should encode, if any claim is to be made that a model has any degree of semantic meaning. However, emotion is not just another modality like vision through a camera or haptic sensations through a robotic hand; emotion is communicative on its own, albeit with limited (but important) social signals, pre-linguistic in that it helps scaffold the language learning process especially early on, and emotion is later intertwined with cognitive development and abstract linguistic meaning. 

How, then, could we represent affect and/or emotion computationally? As has been the case with deriving meaning from text, we can't also derive emotion from text. Pulvermüller \citep{Moseley2012-av,Dreyer2018-fc} I think gives us a hint: the only way we can arrive at a representation of emotion that we could possibly make use of computationally is if we tie emotion to behavior, which is how affect and emotion are signalled between humans. That means we need something to produce that behavior---we've already established that embodiment and interaction are crucial, and that robots are the only computational devices that fulfill the requirements of embodiment because they can act in the physical world. Thus we need humans to watch robots and record their appraisals of robot behaviors for emotional content, then link the behaviors to the emotion. That's only an approximation of emotion through the back door, but it's a start. 

\section{Meaning and the Brain}

Explaining what was missing from computational models of language (like large language models) was easy when I explained the difference between concrete and abstract words \citep{Kennington2022-db}. It's no dichotomy either; some concepts are very concrete in that they exist physically (\emph{chair}), a bit less concrete in that they exist physically but also have abstract properties that make them what they are (\emph{farm}, \emph{city}), and some words are abstract in that there is no physical denotation where the meanings are built upon meanings of other concepts (\emph{democracy}). 

From my readings above, learning that concreteness and abstractness play with emotion in different ways was additional evidence that the concreteness/abstractness dimension of language was something worth my attention. Neuroscience literature further showed that abstract and concrete concepts have different represntational frameworks in the brain \citep{Crutch2005-uf}. However, I found neuroscience literature difficult to digest because a lot of the terminology. 

A book by Iain McGilchrist helped me grasp some of the neuroscience terminology \citep{McGilchrist2019-zt}, and I found that it fed my obsession for the concrete-abstract dimension of language. The main thesis is that the left and right brain hemispheres are, while very similar in function, have some notable differences with many important implicationss. Of course, I was primarily interested in how those differences might affect the language acquisition process. The following largely deal with the concrete-abstract nature of the hemispheres and the role emotion plays:
\begin{itemize}
   
 \item The left hemisphere is the hemisphere of abstraction, which, as the word itself tells us, is the process of wresting things from their context. Thus the right hemisphere does have a vocabulary: it certainly has a lexicon of concrete nouns and imageable words which it shares with the left hemisphere; but, more than that, perceptual links between words are made primarily by the right hemisphere (p.50). In general, then, the left hemisphere’s tendency is to classify, where the right hemisphere’s is to identify individuals (p.52). It has been suggested that our concepts are determined by the language that we speak (the Sapir–Whorf hypothesis). However, this is no more than a half or quarter truth. Children certainly often get the concept first and then quickly learn the word to describe it, which is the wrong way round from the Sapir–Whorf point of view. Moreover there is evidence that five-month-old babies have a concept, to do with tightness of fit, which they subsequently lose if their native language does not embody the same concept (p.110).

 \item ... the right hemisphere’s interest in language lies in all the things that help to take it beyond the limiting effects of denotation to connotation: it acknowledges the importance of ambiguity. It therefore is virtually silent, relatively shifting and uncertain, where the left hemisphere, by contrast, may be unreasonably, even stubbornly, convinced of its own correctness (p.80). 

 \item `emotion binds together virtually every type of information the brain can encode... [it is] part of the glue that holds the whole system together’ (p.88; quoting Douglas Watt)

 \item To recapitulate, then: language originates as an embodied expression of emotion, that is communicated by one individual `inhabiting’ the body, and therefore the emotional world, of another; a bodily skill, further, that is acquired by each of us through imitation, by the emotional identification and intuitive harmonisation of the bodily states of the one who learns with the one from whom it is learnt; a skill moreover that originates in the brain as an analogue of bodily movement, and involves the same processes, and even the same brain areas, as certain highly expressive gestures, as well as involving neurones (mirror neurones) that are activated equally when we carry out an action and when we see another carry it out (so that in the process we can almost literally be said to share one another’s bodily experience and inhabit one another’s bodies) [...] which binds us together as physically embodied beings through a form of extended body language that is emotionally compelling across a large number of individuals within the group (p.122).
\end{itemize}

There are other excerpts from the book that related to language, but these suffice for my purposes here. My primary takeaway is that the concrete-abstract dimension of language is one of the most fundamental aspects of language itself; certainly also of cognition and emotion. In fact, the neurological hardware upon which human thought takes places has, it seems, split the hemispheres to capitalize on the interplay between concreteness and abstractness. Pre-linguistic indeed. As for computational models, large language models like ChatGPT that are trained only on text are purely left-brain models.

\subsection{Implications}

Concreteness and abstractness are well studied in some fields, but taken for granted in NLP research. The concreteness-abstractness divide can help us understand meaning and the assumptions we are making in our models (e.g., that large language models trained on text are purely abstract). Moreover, cognition (which is often equated with language and visa-versa) doesn't stand on its own: cognition needs emotion. 

Incorporating emotion into the language learning process is an additional challenging requirement of putting a computational agent into a setting that is similar to the setting that children are in when they learn their first words. The requirements are, thus far, that the child-like agent must:

\begin{itemize}
    \item be embodied---the agent must be able to act in its environment and potentially manipulate objects and the language model needs access to internal embodied states and sensory modalities of the external world
    \item interact using speech---he agent must use speech as the primary modality for acquiring language, partly because prosody helps carry affective information, and be motivated to interact with others
    \item be physically co-located with language speakers---the agent must be able to visually and auditorially perceive the person(s) that it is learning language from, partly because physical behaviors carry emotional information
    \item distinguish concrete and abstract concepts---be able to learn concrete concepts that denote physical individual things, but also be able to use those concrete concepts abstractly and be able to learn abstract concepts from existing knowledge
    \item use affect and emotion---the agent must use affective displays to facilitate language learning in the early stages, language must ground into affect and emotion concepts are acquired in lock-step with cognitive and abstract language development
\end{itemize}

There are certainly other aspects that I did not explore or find in my years-long search for relevant literature, for example theory-of-mind needs to be modeled and recent work is exploring theory-of-mind deeper, but the entire notion of theory-of-mind needs to be ironed out to the degree that it could be modeled. Likewise, play is an important part of cognitive development in part because it gives children a chance to enact meaning with their bodies and make mistakes with language as they are learning it, and to discover affordances of objects in the world. But I think that theory-of-mind, affordance, and play are, like language, intertwined with emotion.

\paragraph{Acknowledgements} I would like to thank Patty Kennington-Rooks and Vanessa Christensen for helpful and detailed feedback, as well as fruitful discussion. I would also like to thank members of the \emph{Speech, Language, and Interactive Machines} research group at Boise State University for helping to fine-tune some of the ideas. 

\bibliography{paperpile}

\end{document}